\documentclass[a4paper,10pt, conference]{IEEEtran}

\IEEEoverridecommandlockouts

\usepackage[a4paper, left=54pt, right=37pt, top=122pt, bottom=54pt]{geometry}
\newgeometry{left=54pt, right=37pt, top=122pt, bottom=54pt}

\usepackage{cite}
\usepackage{amsmath, amssymb, amsfonts}
\usepackage{algorithmic}
\usepackage{graphicx}
\usepackage{tikz}
\usepackage{textcomp}
\usepackage{multirow}

\usepackage{subcaption}
\usepackage{lipsum}
\usepackage[utf8]{inputenc}

\begin{document}

\title{Attention-Guided Integration of CLIP and SAM for Precise Object Masking in Robotic Manipulation\\
}

\author{\IEEEauthorblockN{1\textsuperscript{st} Muhammad A. Muttaqien\IEEEauthorrefmark{1}\IEEEauthorrefmark{2}}
\IEEEauthorblockA{\textit{Automation Research Team} \\
\textit{National Institute of AIST}\\
Tokyo, Japan \\
muha.muttaqien@aist.go.jp}
\and
\IEEEauthorblockN{2\textsuperscript{nd} Tomohiro Motoda}
\IEEEauthorblockA{\textit{Automation Research Team} \\
\textit{National Institute of AIST}\\
Tokyo, Japan \\
tomohiro.motoda@aist.go.jp}
\and
\IEEEauthorblockN{3\textsuperscript{rd} Ryo Hanai}
\IEEEauthorblockA{\textit{Automation Research Team} \\
\textit{National Institute of AIST}\\
Tokyo, Japan \\
ryo.hanai@aist.go.jp}
\and
\IEEEauthorblockN{4\textsuperscript{th} Domae Yukiyasu}
\IEEEauthorblockA{\textit{Automation Research Team} \\
\textit{National Institute of AIST}\\
Tokyo, Japan \\
domae.yukiyasu@aist.go.jp}
}

\maketitle

\footnotetext[1]{Also affiliated with the Department of Computer Science, University of Tsukuba, Tsukuba, Japan.}

\begin{abstract}
This paper introduces a novel pipeline to enhance the precision of object masking for robotic manipulation within the specific domain of masking products in convenience stores. The approach integrates two advanced AI models, CLIP and SAM, focusing on their synergistic combination and the effective use of multimodal data (image and text). Emphasis is placed on utilizing gradient-based attention mechanisms and customized datasets to fine-tune performance. While CLIP, SAM, and Grad-CAM are established components, their integration within this structured pipeline represents a significant contribution to the field. The resulting segmented masks, generated through this combined approach, can be effectively utilized as inputs for robotic systems, enabling more precise and adaptive object manipulation in the context of convenience store products.
\end{abstract}

\begin{IEEEkeywords}
System Integration, Multimodal Data, Object Masking, Robotic Manipulation
\end{IEEEkeywords}

\section{Introduction}
\begin{figure}[tbp]
    \centering
    \begin{tikzpicture}
        \node[inner sep=0pt, rounded corners=20pt] {\includegraphics[width=\linewidth]{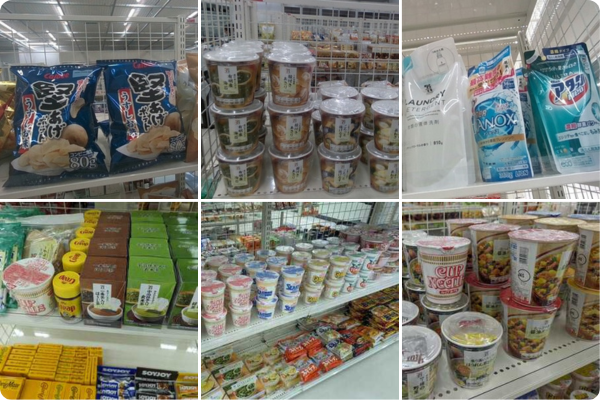}};
    \end{tikzpicture}
    \caption{A convenience store scene displaying various products used for segmentation tasks in our laboratory.}
    \label{fig:cstore}
    \vspace{-5mm}
\end{figure}

In recent years, the ability to recognize and manipulate specific objects within well-defined domains, such as products in convenience stores, has become increasingly important in the field of robotic manipulation \cite{leitner2016acrv} \cite{costanzo2020manipulation} \cite{grotz2023picking}. As robots are expected to perform more complex tasks in diverse environments, the need for precise object identification and interaction grows, particularly in domains where a high level of accuracy is crucial. For instance, in convenience stores (Figure \ref{fig:cstore}), robots must reliably identify and handle a wide variety of products, each with unique visual characteristics, to automate tasks such as stocking, sorting, and customer assistance.

The advent of large foundation models like CLIP (Contrastive Language-Image Pretraining) \cite{radford2021learning} and SAM (Segment Anything Model)\cite{kirillov2023segment} has opened new possibilities for improving object recognition and segmentation tasks. These foundation models, which are pre-trained on vast datasets, provide a robust framework for tackling complex AI challenges. CLIP’s ability to align images and text through a shared embedding space, combined with SAM’s flexibility in segmenting any part of an image, offers a strong foundation for developing advanced robotic systems. Fine-tuning CLIP for specific applications within a system can enhance the precision of the prompts that guide SAM in producing accurate object masks, which are critical for robots to perform complex manipulation tasks more accurately and reliably.

However, integrating language and vision in multi-modal tasks presents significant challenges. While models like CLIP and SAM are powerful individually, their real-world use is limited by a key bottleneck: the lack of domain-specific data, such as object detection, grounding, and caption data, which are essential for fine-tuning. Existing pipelines like GroundedSAM\cite{ren2024grounded}, FastSAM\cite{zhao2023fast}, or even YOLO World\cite{redmon2023yoloworld} integrated with SAM perform well in generalized settings but struggle in highly specific environments, such as convenience stores, where the variety of objects and their contextual relevance differ significantly from the data these models were trained on. The difficulty lies in the fact that preparing such domain  time-consuming and labor-intensive.

In this paper, we propose a novel pipeline that integrates fine-tuned CLIP and SAM, enhanced by gradient-based attention mechanisms, to address the challenges of object masking in robotic manipulation. Focusing on convenience stores, our method optimizes performance with customized datasets and task-specific fine-tuning, overcoming the data limitations faced by existing pipelines. This allows for more precise and reliable robotic systems in real-world applications. The remainder of this paper discusses our approach, the challenges overcome, and the potential applications for enhancing robotic systems.

\begin{figure*}[tbp]
  \centering
  \includegraphics[width=\linewidth]{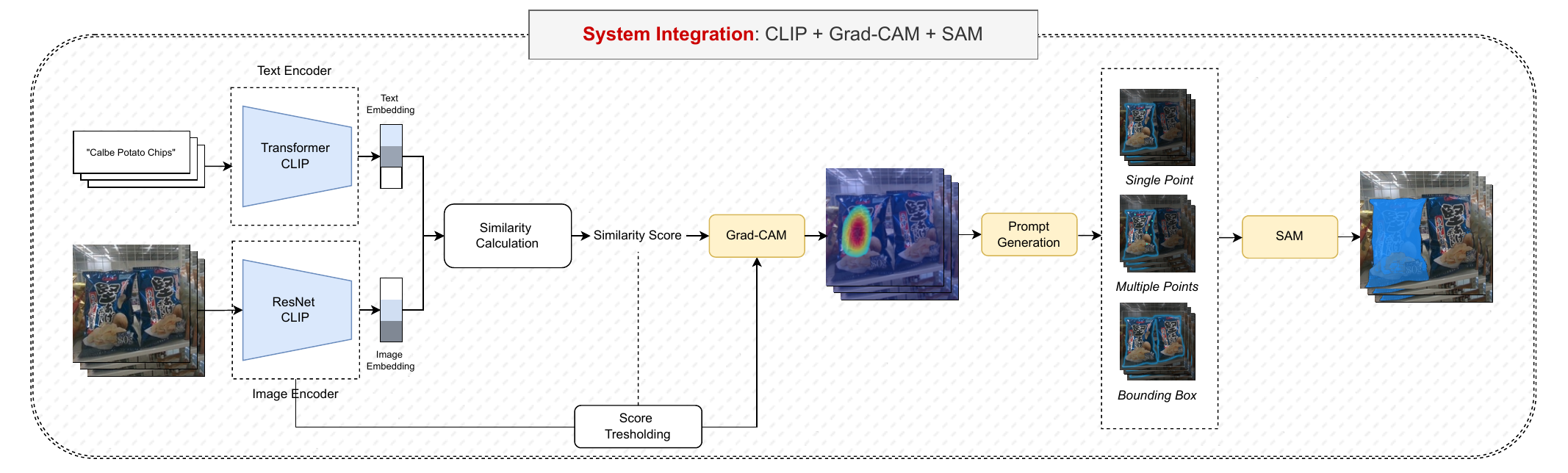}
  \caption{Summary of our system integration. The diagram illustrates the flow from input processing in CLIP to the generation of accurate segmentation masks by SAM, guided by Grad-CAM attention maps.}
  \label{fig:pipeline}
  \vspace{-2mm}
\end{figure*}

\section{Related Work}

The development of object recognition, segmentation, and multimodal AI has been significantly shaped by pioneering models like AlexNet \cite{krizhevsky2012imagenet}, VGGNet \cite{simonyan2014very} and ResNet \cite{he2016deep}, which introduced deep architectures capable of capturing intricate visual patterns. VGGNet, known for its deep and simple structure, uses small convolution filters to achieve high accuracy, while ResNet introduced residual learning, enabling deeper networks without the vanishing gradient problem. These advancements in convolutional neural networks (CNNs) have laid the groundwork for more sophisticated models that are crucial for tasks such as object manipulation in complex environments like convenience stores.

Object detection has also evolved with the introduction of models such as Faster R-CNN \cite{ren2015faster} and YOLO \cite{redmon2016you}, both of which have significantly influenced the field. Faster R-CNN integrates region proposal networks with CNNs to enhance detection accuracy, making it a preferred choice for tasks requiring precision. In contrast, YOLO redefined object detection by framing it as a single regression problem, allowing for real-time performance. The progression of object detection models has been further supported by attention mechanisms, particularly through the Transformer architecture and its adaptation to visual tasks with Vision Transformers (ViT) \cite{dosovitskiy2021image}. ViT demonstrated that self-attention could effectively capture global dependencies in images, offering a new approach to image recognition that reduces reliance on convolutional operations.

The integration of language and vision has opened new possibilities in multimodal AI, with models like CLIP, GLIP 
\cite{yu2022glip}, and DINO \cite{caron2021emerging} playing a central role. CLIP, developed by OpenAI, showed the power of aligning text and image representations, enabling robust zero-shot learning across various vision tasks. Building on this, GLIP enhanced the grounding of language in visual contexts, making it more effective for object detection in multimodal scenarios. DINO and its extension, Grounding DINO \cite{li2023grounding}, explored self-supervised learning for visual representations, emphasizing the importance of grounding visual concepts in multimodal inputs. These advancements have significantly enhanced the capabilities of systems that require a deep understanding of both visual and textual information.

Segmentation, a crucial aspect of object manipulation, has advanced significantly with models like the SAM by Meta AI. SAM represents a major leap in segmentation technology, delivering high-quality results across a wide range of objects and scenes. Trained using a vast and diverse dataset, SAM enhances its ability to generalize across various environments. These advanced segmentation models are particularly relevant in environments like convenience stores, where precise object detection and manipulation are critical.

To ensure the reliability and interpretability of these complex systems, techniques like Grad-CAM \cite{selvaraju2017grad} have been developed to provide visual explanations of model decisions. Grad-CAM highlights the regions of an image that influence a model’s output, offering insights into the model’s decision-making process. This explainability is essential in multimodal AI systems, where understanding the model’s focus areas can lead to more trustworthy and effective applications.

\section{Foundation Model}

Foundation models have revolutionized artificial intelligence by providing versatile, pre-trained models that can be adapted across various tasks and domains. These models, trained on extensive datasets, capture a wide range of patterns and representations, making them invaluable for complex applications like robotic manipulation. Their scalability and ability to be fine-tuned for specific tasks allow for significant advancements in performance, particularly in environments requiring precise and reliable AI systems.

CLIP (Contrastive Language-Image Pretraining) is a prime example of a foundation model that bridges the gap between language and vision. Developed by OpenAI, CLIP uses a dual-encoder architecture to align images and text within a shared embedding space. This capability enables CLIP to perform zero-shot learning and effectively recognize and distinguish objects in varied contexts. In our research, CLIP is crucial for generating accurate prompts that guide the segmentation process, particularly in the complex and diverse environment of convenience stores.

SAM (Segment Anything Model) complements CLIP by offering high-precision image segmentation. Built on a vision transformer architecture, SAM processes images to create detailed segmentation masks based on user-defined prompts. Its versatility and accuracy are crucial for robotic manipulation, particularly in complex environments like convenience stores, where detecting objects on cluttered shelves can be challenging. By integrating SAM with CLIP, we achieve a more precise and adaptive object masking process, enhancing the ability of robots to interact accurately with products in such demanding settings.

The relevance of these foundation models to our research lies in their ability to tackle the challenges of multi-modal AI tasks. As shown in Figure \ref{fig:pipeline}, the integration of CLIP and SAM, enhanced by gradient-based attention mechanisms like a modified Grad-CAM, forms the core of our proposed pipeline. By fine-tuning on domain-specific datasets, this approach enhances the system's ability to accurately recognize and mask objects of interest in specific domains, such as convenience stores, based on defined prompts.

\section{Gradient-based Attention}

In this section, we discuss the role of gradient-based attention mechanisms in our system integration, focusing on how Grad-CAM is utilized to enhance the interpretability and effectiveness of the CLIP and SAM models within our proposed pipeline. We will explore two key applications: Guided Grad-CAM and its use in Image Captioning models.

\subsection{Guided Grad-CAM}
Guided Grad-CAM is a technique that combines the strengths of two visualization methods: Grad-CAM and Guided Backpropagation. Grad-CAM generates class-discriminative localization maps, highlighting the important regions in an image that contribute to a model’s decision. However, while Grad-CAM identifies relevant regions, it lacks fine-grained detail. Guided Backpropagation offers high-resolution visualizations but lacks class discrimination. Combining these methods, Guided Grad-CAM offers high-resolution, class-discriminative visual explanations, enhancing the understanding and refinement of attention mechanisms.

\subsection{Image Captioning Model Using Grad-CAM}
Grad-CAM is also evident to be effectively applied to image captioning models to understand how visual features are linked to specific words in a generated caption. In this context, Grad-CAM helps to identify the regions of an image that are most influential in predicting each word of the caption, providing insights into the model's decision-making process. This capability is particularly useful when integrating the outputs of CLIP with SAM. By using Grad-CAM to analyze the attention maps generated during image captioning, we can better align the visual and textual information processed by CLIP.

The integration of Grad-CAM within our system enhances both the interpretability and precision of the CLIP and SAM models. By employing Guided Grad-CAM and leveraging its application in image captioning, we ensure that the visual attention mechanisms are aligned with the task-specific goals of our pipeline, ultimately leading to more accurate and adaptive object masking for robotic manipulation. This approach is particularly effective in the convenience store environment we focus on.

\section{Technical Setup}

\subsection{Datasets}

To fine-tune the CLIP model for our application, we prepared a dataset of approximately 30,000 high-resolution RGB images paired with text (without the need for detection or grounding data) from a realistic mock-up of a convenience store in our laboratory. The dataset includes a wide variety of products across multiple categories, capturing diverse angles, lighting conditions, and arrangements. While the dataset contains many specific items, we highlight some specific products for detailed visualization.

\subsection{CLIP: Object Recognition}

CLIP employs two primary architecture variants: ViT (Vision Transformer) and ResNet-based models. For our implementation, we focused on the ResNet variant because its architecture is more suitable for gradient-based attention techniques, such as Grad-CAM, which we use not only to enhance the interpretability of the model but also to effectively connect CLIP and SAM in our pipeline. The ResNet-based CLIP model processes images through a series of convolutional layers, which are followed by fully connected layers that map the visual data into a shared embedding space with text.

For fine-tuning the ResNet-based CLIP model, we selected hyperparameters optimized for our domain-specific dataset, as depicted in Table \ref{tab:hyperparameters}. These settings were chosen based on preliminary experiments to balance training efficiency and model performance. The ResNet-based CLIP model, with approximately 150 million parameters, captures complex visual patterns and effectively aligns them with corresponding textual descriptions, seamlessly integrating with SAM for precise object masking in our system.

\begin{table}[h!]
\centering
\caption{Hyperparameters used in the CLIP model training.}
\begin{tabular}{|p{4cm}|p{2cm}|}
\hline
\textbf{Hyperparameters} & \textbf{Value} \\ [1.5ex] \hline
Batch Size             & 64            \\ [0.75ex] 
Learning Rate          & 1e-5          \\ [0.75ex] 
Number of Epochs       & 50            \\ [0.75ex] 
Optimizer              & Adam          \\ [0.75ex] 
Betas                  & (0.9, 0.98)   \\ [0.75ex] 
Weight Decay           & 0.01          \\ [0.75ex] 
Input Size (Image)     & 224x224       \\ [0.75ex] \hline
\end{tabular}
\label{tab:hyperparameters}
\vspace{-2mm}
\end{table}

\subsection{SAM: Object Masking}

SAM is available in several variants, including base, large, and huge, each differing in model size, number of parameters, and computational complexity. The base variant is lightweight and suitable for limited resources, while the huge variant offers the highest accuracy but requires more computational power. The large variant achieves a balance between these extremes, offering a good compromise between performance and efficiency. The large SAM model has about 640 million parameters, making it more powerful than the base model's 300 million parameters but less demanding than the huge variant, which has over 1 billion. Additionally, we utilized SAM's promptable features, experimenting with single-point, multiple-point, and bounding box prompts to further refine the segmentation process.

\subsection{Grad-CAM}

Grad-CAM was chosen for its ability to work without requiring changes to the CLIP architecture and its proven effectiveness with multimodal inputs like image captioning and visual question answering. We explore how Grad-CAM is applied in our system to enhance the effectiveness of the SAM model using different types of prompts: Single Point, Multiple Points, and Bounding Box.

\begin{itemize}

\item Single Point: Grad-CAM generates attention maps by focusing on a single pixel, providing localized insight into the model's focus and refining segmentation masks around specific points of interest. This is especially useful for precise segmentation of small or detailed objects.

\item Multiple Points: Applying Grad-CAM to multiple points on the object of interest creates a more comprehensive attention map, covering different areas of the object. This method improves the model's ability to accurately segment complex shapes or objects with varying features by integrating information from key locations.

\item Bounding Box: Grad-CAM is also used with bounding box prompts, where the model focuses on a rectangular region surrounding the object. This approach is effective for segmenting larger objects or when the general area of the object is known. The bounding box prompt guides the segmentation process, ensuring that attention remains within the specified boundaries.

\end{itemize}

\subsection{System Integration}

We outline a pipeline integrating CLIP, SAM, and Grad-CAM to achieve precise object masking for robotic manipulation in convenience stores. The process begins with CLIP, which calculates a similarity score between the image and the prompt. A high similarity score suggests the presence of the object, which is then visualized using Grad-CAM through the ResNet-based model in CLIP. Grad-CAM generates an attention map that highlights relevant areas of the image. This attention map is then converted into prompts—single points, multiple points, or bounding boxes—that guide SAM to accurately mask the object. As illustrated in Figure \ref{fig:pipeline}, this pipeline demonstrates how CLIP’s output informs SAM, with SAM optimizing the process for real-time applications.

To determine the optimal threshold for our application, we calculate similarity scores between each product image and all product names using the fine-tuned CLIP model. Each image is classified as correct if the highest similarity score matches the correct product name; otherwise, it is classified as incorrect. By analyzing these scores, we compute the F1-score at various thresholds through precision-recall calculations. The threshold that maximizes the F1-score provides the best balance between precision and recall, indicating when a similarity score is sufficiently high to classify an object as present. In this case, the optimal score threshold is 0.489.

The process starts with CLIP’s visual encoding of the input image \( I \) using a visual encoder \( f_v \), producing the visual embedding \( V \). Simultaneously, the text prompt \( t \) is processed by the text encoder \( f_t \), resulting in the text embedding \( T \). These embeddings are normalized to \( E_V \) and \( E_T \), respectively. The dot product of these normalized embeddings yields the similarity score \( S \), indicating the relevance of the object in the image.

\begin{equation}
\begin{aligned}
& \quad V = f_v(I) \\
& \quad T = f_t(t) \\
& \quad E_V = \frac{V}{\|V\|} \\
& \quad E_T = \frac{T}{\|T\|} \\
& \quad S = E_V \cdot E_T \\
& \quad A(x, y) = \text{ReLU}\left(\sum_k \alpha_k \cdot f_k(x, y)\right) \\
& \quad P = f(A) \in \{\text{single point, multiple points, bounding box}\} \\
& \quad M = g(I, P) \\
\end{aligned}
\end{equation}

Next, Grad-CAM generates an attention map \( A(x, y) \) based on the similarity score, highlighting relevant areas in the image. Here, \(x\) and \(y\) represent the spatial coordinates of a pixel within the attention map. This attention map is then converted into prompts \( P \) (such as single points, multiple points, or bounding boxes that guide the SAM model). Finally, SAM uses the image and prompts to generate the segmentation mask \( M \), accurately isolating the object of interest. This pipeline seamlessly integrates CLIP’s multimodal capabilities with SAM’s precise segmentation, enhanced by Grad-CAM’s attention mapping.

\section{Experiments}

In this section, we evaluate the performance of our proposed pipeline, focusing on the effectiveness of our approach in improving object recognition, segmentation accuracy, and overall system reliability. We also discuss the insights gained from our analysis and the technical limitations that impact the system's applicability in real-world scenarios.

\subsection{CLIP Model Fine-Tuning and Performance Evaluation}

We implemented our pipeline using Python programming and the PyTorch Deep Learning Framework \cite{paszke2019pytorch}, and trained the models on an NVIDIA GeForce RTX 4090. Before fine-tuning the CLIP model, we assessed its original performance on our domain-specific dataset, which focuses on convenience store products. We used zero-shot and top-5 accuracy metrics to evaluate how well CLIP could recognize and distinguish similar products without additional training. We initially used CLIP as a classifier to measure these metrics and then compared its performance with that of fine-tuned versions using ResNet and ViT backbones.

\begin{table}[h!]
\centering
\caption{Comparison of Accuracy for Different CLIP Models}
\begin{tabular}{|l|l|c|c|}
\hline
\textbf{Method} & \textbf{Type} & \textbf{Zero-shot Accuracy} & \textbf{Top-5 Accuracy} \\ [1.5ex] \hline
\multirow{2}{*}{Original} & Product & 0.0942 & 0.2586 \\ [0.75ex] 
                               & Category & 0.0796 & 0.3047 \\ [0.75ex] \hline
\multirow{2}{*}{ResNet50} & Product  & 0.5372 & 0.8168 \\ [0.75ex] 
                               & Category & 0.1749 & 0.3225 \\ [0.75ex] \hline
\multirow{2}{*}{ResNet101} & Product  & 0.3770 & 0.7351 \\ [0.75ex] 
                               & Category & 0.1550 & 0.3120 \\ [0.75ex] \hline
\multirow{2}{*}{ViT16}    & Product  & 0.8000 & 0.9539 \\ [0.75ex] 
                               & Category & 0.2513 & 0.3812 \\ [0.75ex] \hline
\multirow{2}{*}{ViT32}    & Product  & 0.5319 & 0.8356 \\ [0.75ex] 
                               & Category & 0.1654 & 0.3435 \\ [0.75ex] \hline
                               
\end{tabular}
\label{tab:clipaccuracy}
\vspace{-2mm}
\end{table}

As shown in Table \ref{tab:clipaccuracy}, the original CLIP model struggles to recognize products in our specific domain, which is understandable given that many of the products have Japanese names, such as Edamariko and Tonkotsu. In this context, "Product" refers to specific items like Edamariko or Tonkotsu, while "Category" refers to broader classifications such as snacks. Both ResNet and ViT-based CLIP models show significant improvement after being trained on our prepared datasets. Interestingly, smaller models like ResNet50 and ViT16 outperform their larger counterparts, such as ResNet101 and ViT32. This suggests that the larger models may be prone to overfitting, whereas the smaller models generalize better for the task at hand.

\begin{figure}[h!]
    \centering
    \begin{subfigure}[b]{0.45\linewidth}  
        \centering
        \includegraphics[width=\linewidth]{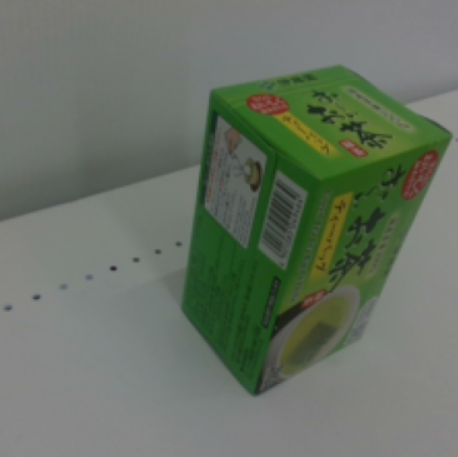}
        \caption{Oi Ocha Tea Bag. Top predictions: Oi Ocha Tea Bag (98.83\%), Kleenex Lotion Tissue Soft Pack (0.29\%), House Cream Stew Origin (0.17\%).}
    \end{subfigure}
    \hfill
    \begin{subfigure}[b]{0.45\linewidth}  
        \centering
        \includegraphics[width=\linewidth]{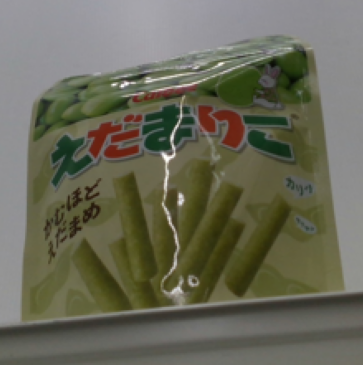}
        \caption{Edamariko. Top prediction: SEVEN\&i PREMIUM Stockings Nude Beige (28.88\%), Edamariko (23.57\%), House Cream Stew Origin (11.49\%).}
    \end{subfigure}

    \vskip\baselineskip  

    \begin{subfigure}[b]{0.45\linewidth}  
        \centering
        \includegraphics[width=\linewidth]{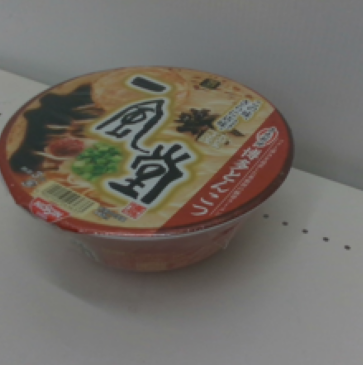}
        \caption{Shinaji Hakata Tonkotsu. Top predictions: Shinaji Hakata Tonkotsu (77.59\%), Maruchan Midorinotanuki (13.07\%), Edamariko (2.53\%).}
    \end{subfigure}
    \hfill
    \begin{subfigure}[b]{0.45\linewidth}  
        \centering
        \includegraphics[width=\linewidth]{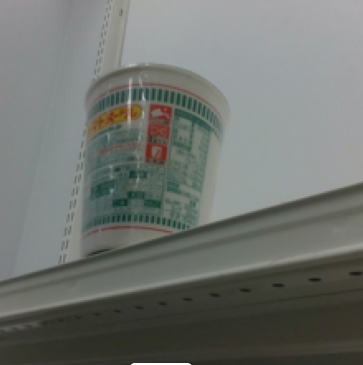}
        \caption{Chili Tomato Noodle. Top predictions: Chili Tomato Noodle (85.69\%), Maruchan Midorinotanuki (11.60\%), House Cream Stew Origin (1.11\%).}
    \end{subfigure}

    \caption{Top predictions and confidence scores for four different images. Each image is shown with a caption listing the top predictions generated by the model.}
    \label{fig:similarity4}
    \vspace{-2mm}
\end{figure}

After establishing baseline performance, we shifted to treating CLIP as a model capable of calculating the similarity between image captions and image inputs. This approach leveraged CLIP’s multimodal capabilities, where it processes text and image inputs to produce a similarity score. Figure \ref{fig:similarity4} illustrates that a high score indicates the presence of the object of interest in the image. The fine-tuned CLIP model can adapt to our specific dataset, improving its ability to accurately recognize and match products with their descriptions.

\subsection{Attention Map Analysis}

\begin{figure}[h!]
    \centering
    \begin{subfigure}[b]{0.30\linewidth}
        \centering
        \includegraphics[width=\linewidth]{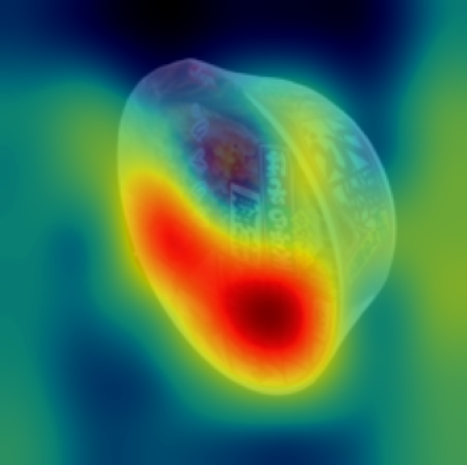}
        \caption{Top Part}
    \end{subfigure}
    \hfill
    \begin{subfigure}[b]{0.30\linewidth}
        \centering
        \includegraphics[width=\linewidth]{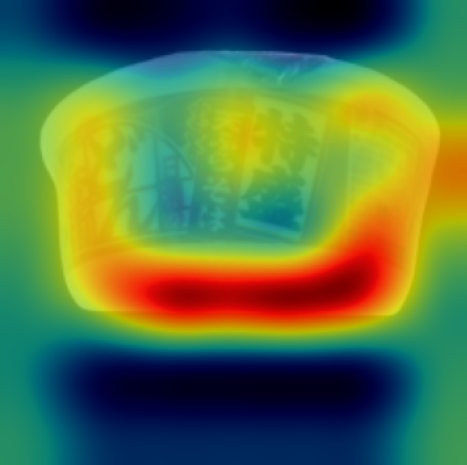}
        \caption{Bottom Part}
    \end{subfigure}
    \hfill
    \begin{subfigure}[b]{0.30\linewidth}
        \centering
        \includegraphics[width=\linewidth]{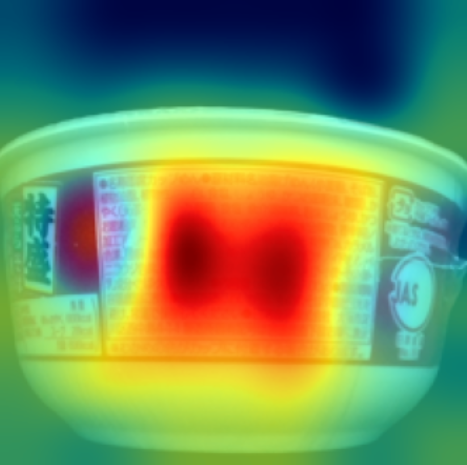}
        \caption{Side Part}
    \end{subfigure}

    \caption{Attention maps highlighting different parts of the object: (a) Top Part, (b) Bottom Part, and (c) Side Part. These visualizations show where the CLIP model focuses, providing crucial guidance for the segmentation process.}
    \label{fig:attention}
    \vspace{-4mm}
\end{figure}

We generated attention maps using Grad-CAM applied to the fine-tuned CLIP model. These maps visually represent the areas of an image that the model focuses on when making decisions. To evaluate the performance of these attention maps, we visually analyzed whether the highlighted regions were located within the object of interest. If the attention was focused inside the object of interest, we considered the map effective, regardless of the specific part of the object being highlighted as shown in Figure \ref{fig:attention}. This evaluation was key in assessing how effectively the fine-tuned model focuses on relevant regions, validating the fine-tuning success. The attention maps also laid the groundwork for integrating with the SAM model, ensuring accurate segmentation masks.

\subsection{Attention-Guided Integration}

\begin{figure}[h!]
    \centering
    \begin{subfigure}[b]{0.30\linewidth}
        \centering
        \includegraphics[width=\linewidth]{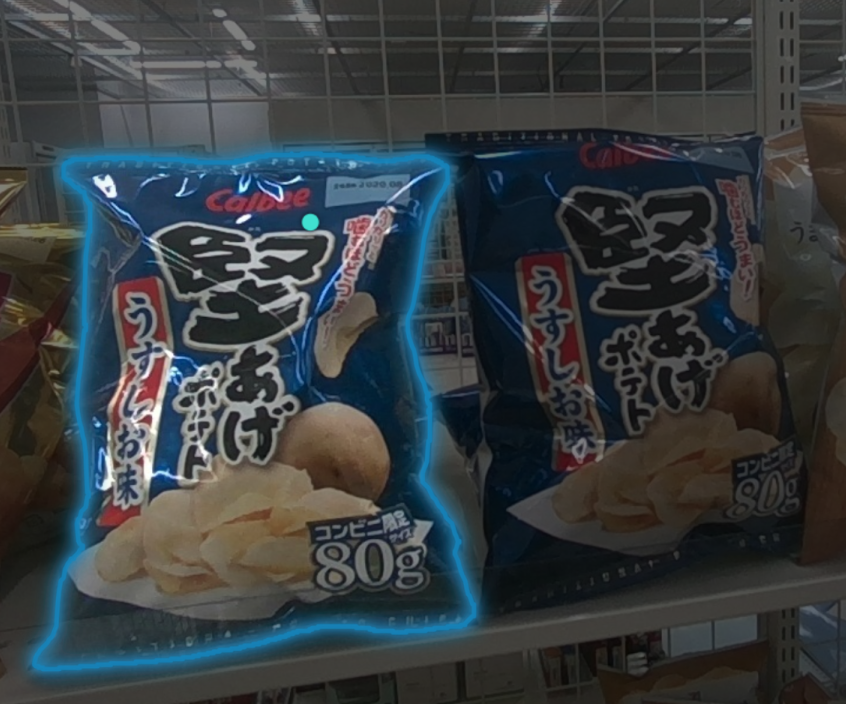}
        \caption{Single Point}
    \end{subfigure}
    \hfill
    \begin{subfigure}[b]{0.30\linewidth}
        \centering
        \includegraphics[width=\linewidth]{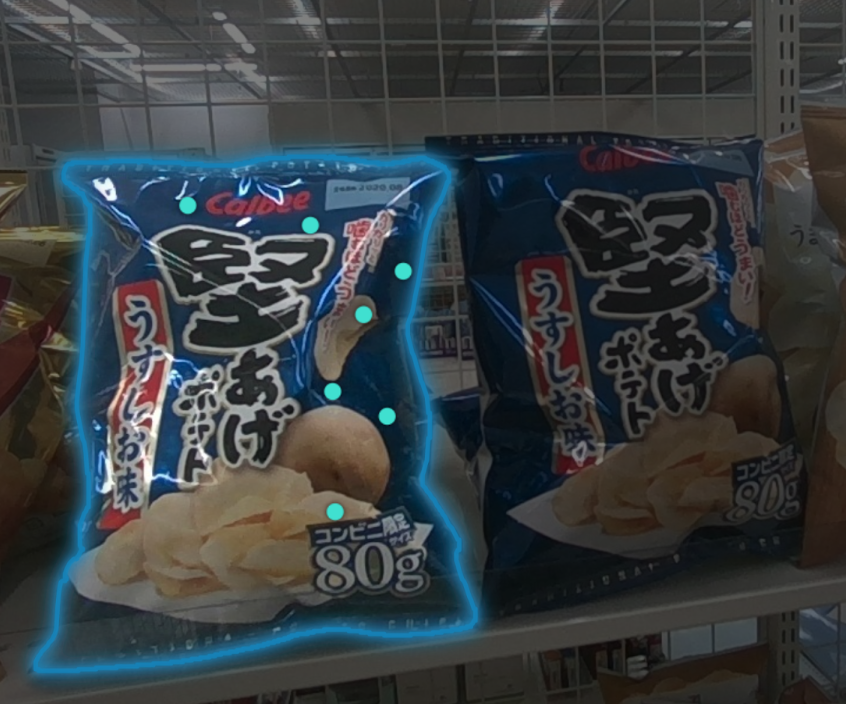}
        \caption{Multiple Points}
    \end{subfigure}
    \hfill
    \begin{subfigure}[b]{0.30\linewidth}
        \centering
        \includegraphics[width=\linewidth]{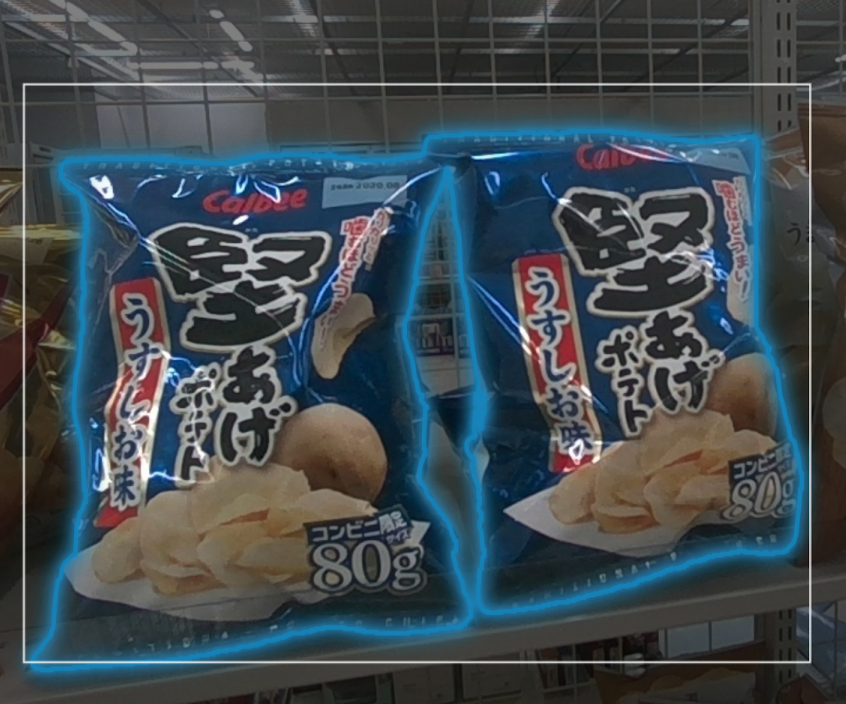}
        \caption{Bounding Box}
    \end{subfigure}
    
    \vspace{4mm} 
    
    \begin{subfigure}[b]{0.30\linewidth}
        \centering
        \includegraphics[width=\linewidth]{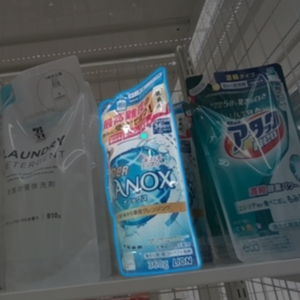}
        \caption{Single Point}
    \end{subfigure}
    \hfill
    \begin{subfigure}[b]{0.30\linewidth}
        \centering
        \includegraphics[width=\linewidth]{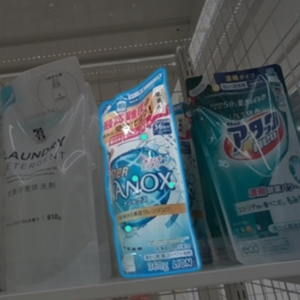}
        \caption{Multiple Points}
    \end{subfigure}
    \hfill
    \begin{subfigure}[b]{0.30\linewidth}
        \centering
        \includegraphics[width=\linewidth]{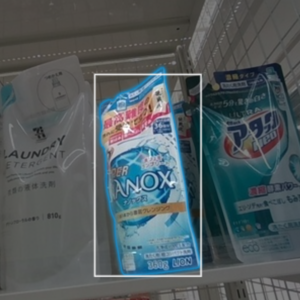}
        \caption{Bounding Box}
    \end{subfigure}

    \caption{Different segmentation prompts: (a) Single Point, (b) Multiple Points, and (c) Bounding Box, each guide the segmentation process in the SAM model to accurately identify objects.}
    \label{fig:samprompts}
    \vspace{-4mm}
\end{figure}

The integration of CLIP and SAM in our pipeline is enhanced by the use of attention maps generated by Grad-CAM. These maps serve as a guide for the SAM model, helping it produce more precise and accurate segmentation masks. The attention maps highlight the most critical areas, with the most red regions indicating the most distinctive features of the object. These regions are then treated as key prompts for SAM, which we assume can leverage these areas, and then serve as a strong prompt for SAM  to create the final segmentation masks. By converting these highly focused areas into prompts (single point, multiple points or bounding box as illustrated in Figure \ref{fig:samprompts}), the SAM model is better equipped to generate accurate object masks in various scenarios.

\begin{figure*}[tbp]
    \centering
    \begin{subfigure}[b]{0.32\linewidth}
        \includegraphics[width=\linewidth]{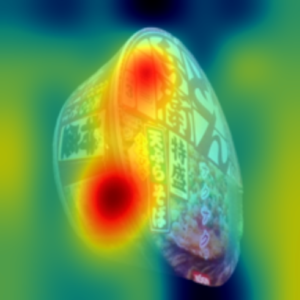}
        \caption{Attention Map}
    \end{subfigure}
    \begin{subfigure}[b]{0.32\linewidth}
        \includegraphics[width=\linewidth]{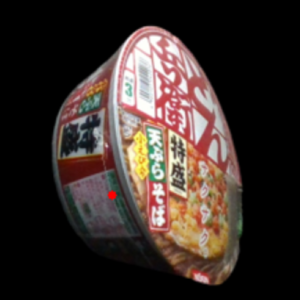}
        \caption{Single Point}
    \end{subfigure}
    \begin{subfigure}[b]{0.32\linewidth}
        \includegraphics[width=\linewidth]{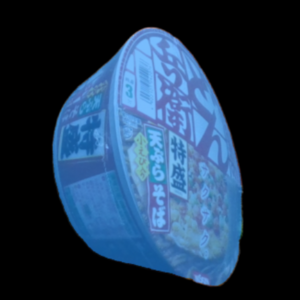}
        \caption{Masked Image}
    \end{subfigure}

    \begin{subfigure}[b]{0.32\linewidth}
        \includegraphics[width=\linewidth]{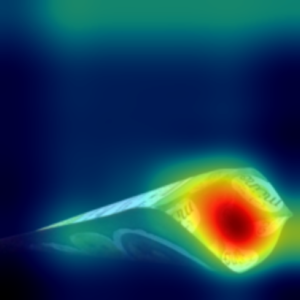}
        \caption{Attention Map}
    \end{subfigure}
    \begin{subfigure}[b]{0.32\linewidth}
        \includegraphics[width=\linewidth]{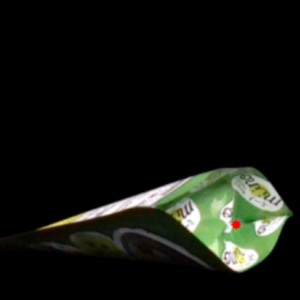}
        \caption{Single Point}
    \end{subfigure}
    \begin{subfigure}[b]{0.32\linewidth}
        \includegraphics[width=\linewidth]{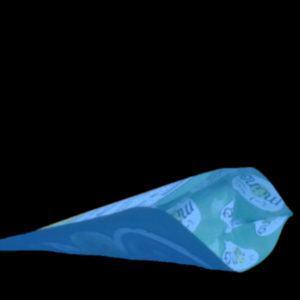}
        \caption{Masked Image}
    \end{subfigure}

    \begin{subfigure}[b]{0.32\linewidth}
        \includegraphics[width=\linewidth]{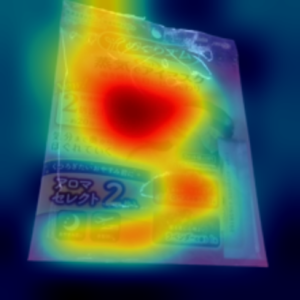}
        \caption{Attention Map}
    \end{subfigure}
    \begin{subfigure}[b]{0.32\linewidth}
        \includegraphics[width=\linewidth]{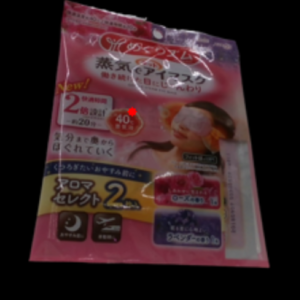}
        \caption{Single Point}
    \end{subfigure}
    \begin{subfigure}[b]{0.32\linewidth}
        \includegraphics[width=\linewidth]{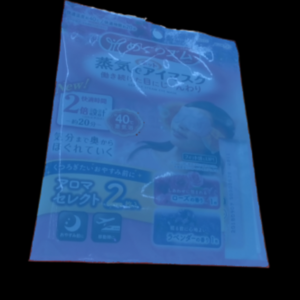}
        \caption{Masked Image}
    \end{subfigure}
    
    \caption{The figure illustrates our system's process of attention-guided object masking across three different items: instant noodles, a snack bag, and a packaged tissue product. For each item, the attention map (a, d, g) highlights key regions that the model focuses on. These regions are then used to create a single-point prompt (b, e, h), which guides the SAM model to accurately generate the final object mask (c, f, i). This sequence demonstrates the system’s ability to effectively segment objects based on distinct visual features in a convenience store setting.}
    \label{fig:maskedimg}
    \vspace{-2mm}
\end{figure*}

The system calculates the coordinates of the red areas in the attention map and converts them into points for the SAM model. If there are multiple red areas, they are transformed into multiple points, allowing for more precise prompts. We found that SAM performs better with multiple points as prompts, as this approach reduces ambiguity about which object is being targeted in the image. If only a single red area is produced by Grad-CAM, the system implements local random sampling by considering a region around the highest activation point. From this local region, 3 points are randomly sampled to serve as input for the SAM model.

Additionally, a bounding box can be generated from the attention map by using the coordinates of the top-left and bottom-right red areas to define the corners of the box. However, we found that this approach is not effective, especially for objects where the attention map covers only a small part of the object of interest. For example, with Calbe Potato Chips, the attention is most likely focused on the brand name on the package, whereas for an apple, the attention is more likely to cover the entire object. Furthermore, a bounding box would be more appropriate if the system needs to detect multiple objects. Therefore, in our case, the bounding box serves as an optional mode within the system.

\subsection{Masking Result}

The integration of CLIP and SAM in our pipeline is significantly enhanced by the use of attention maps generated by Grad-CAM. These maps guide the SAM model in producing more precise and accurate segmentation masks by directing the model's focus to the most relevant areas of the image. To evaluate the effectiveness of this attention-guided integration, we tested how well the attention maps functioned as prompts for SAM, exploring different prompt types, including single point, multiple points, and bounding box. Each prompt type contributed uniquely to the object masking quality, adapting to different scenarios and object complexities.

To assess masking accuracy, we compared the generated masks with ground truth masks using Intersection over Union (IoU) as the evaluation metric, calculating IoU scores by evaluating the predicted masks against the label masks. As shown in Table \ref{tab:iouscores}, ResNet50 CLIP model consistently outperformed the ResNet101 CLIP model across all three prompts demonstrating higher IoU scores. This suggests that the smaller ResNet50 model is better suited to our task, likely due to reduced overfitting. Additionally, the table indicates that using points as prompts for SAM yields better performance than bounding boxes, supporting the effectiveness of point-based prompts in improving mask accuracy.

\begin{table}[h!]
\centering
\caption{IoU Scores for Different Prompt Types}
\begin{tabular}{|c|c|c|}
\hline
\textbf{Method}       & \textbf{Prompts}   & \textbf{IoU Score} \\ [1.5ex] \hline
\multirow{3}{*}{ResNet50 CLIP} & Single Point    & 0.884          \\ [0.75ex] 
                            & Multiple Points  & 0.914           \\ [0.75ex] \hline
\multirow{3}{*}{ResNet101 CLIP} & Single Point    & 0.950          \\ [0.75ex] 
                            & Multiple Points  & 0.957         \\ [0.75ex] 
                            \hline
\end{tabular}
\label{tab:iouscores}
\end{table}

This comparison revealed that the focus areas identified by CLIP were effectively utilized by SAM, resulting in accurate masking across various test cases. The high IoU scores confirmed the robustness and reliability of our masking process, which is essential for ensuring accurate and effective robotic manipulation tasks in real-world applications. Figure \ref{fig:maskedimg} demonstrates how our system highlights key regions in the attention map, selects a single point for segmentation, and accurately masks the identified object.

\subsection{Pipeline Evaluation}

The accuracy results of fine-tuned ResNet-based and ViT-based CLIP models for recognizing products in our specific domain (convenience store) demonstrate the potential of CLIP to be effectively integrated into our pipeline for solving the object recognition problem. Furthermore, by leveraging the gradient-based attention mechanism of Grad-CAM to integrate CLIP with SAM, we have shown the effectiveness of this approach, as evidenced by the remarkable IOU scores achieved with both single and multiple points as prompts.

However, while our system demonstrates strong performance in specific conditions, there are limitations that need to be addressed. The current system is optimized for a controlled camera position, which limits its flexibility in more variable environments. Additionally, the focus on single-object masking restricts the system's ability to handle multiple objects simultaneously.

\section{Conclusion}

In this paper, we have presented a novel pipeline that integrates CLIP, SAM, and Grad-CAM to enhance the precision of object masking for robotic manipulation in convenience stores. By leveraging the strengths of these advanced AI models and fine-tuning them with customized datasets, we have demonstrated that the integration of these components can significantly improve the accuracy and relevance of segmentation tasks. The resulting masks are highly effective as inputs for robotic systems, enabling more accurate manipulation of products in dynamic environments.

Future work will focus on extending the pipeline to handle multiple object masking in complex scenes with overlapping objects. We also aim to improve scalability for a broader range of products and explore other foundation models like YOLO World and GPT \cite{radford2020language} to enhance robustness and versatility. These advancements will make our solution applicable to a wider array of real-world scenarios.

\section*{Acknowledgment}

We would like to express our sincere gratitude to the National Institute of Advanced Industrial Science and Technology (AIST) for their invaluable support and resources that made this research possible. Their contribution was essential in the successful completion of this research work.


\begin{thebibliography}{99}

\bibitem{leitner2016acrv}
J. Leitner, A. W. Tow, N. Sünderhauf, J. E. Dean, J. W. Durham, M. Cooper, M. Eich, C. Lehnert, R. Mangels, C. McCool, P. T. Kujala, L. Nicholson, T. Pham, J. Sergeant, L. Wu, F. Zhang, B. Upcroft, and P. Corke,
"The ACRV Picking Benchmark: A Robotic Shelf Picking Benchmark to Foster Reproducible Research,"
in \textit{Proceedings of the IEEE International Conference on Robotics and Automation (ICRA)}, pp. 1-10, 2016.

\bibitem{costanzo2020manipulation}
M. Costanzo, S. Stelter, C. Natale, S. Pirozzi, G. Bartels, A. Maldonado, and M. Beetz,
"Manipulation Planning and Control for Shelf Replenishment,"
in \textit{IEEE Robotics and Automation Letters}, vol. 1, no. 2, pp. 1-10, 2020.

\bibitem{grotz2023picking}
M. Grotz, S. Atar, Y. Li, P. Torrado, B. Yang, N. Walker, M. Murray, M. Cakmak, and J. R. Smith,
"Towards robustly picking unseen objects from densely packed shelves,"
in \textit{Proceedings of Robotics: Science and Systems (RSS)}, 2023.

\bibitem{radford2021learning}
A. Radford, J. W. Kim, C. Hallacy, A. Ramesh, G. Goh, S. Agarwal, G. Sastry, A. Askell, P. Mishkin, J. Clark, G. Krueger, and I. Sutskever,
"Learning Transferable Visual Models From Natural Language Supervision,"
in \textit{Proceedings of the 38th International Conference on Machine Learning}, PMLR 139:8748-8763, 2021.

\bibitem{kirillov2023segment}
A. Kirillov, E. Mintun, N. Ravi, H. Mao, C. Rolland, L. Gustafson, T. Xiao, S. Whitehead, A. C. Berg, W.-Y. Lo, P. Dollar, and R. Girshick,
"Segment Anything,"
in \textit{Proceedings of the IEEE/CVF International Conference on Computer Vision (ICCV)}, 2023, pp. 4015-4026.

\bibitem{ren2024grounded} T. Ren, S. Liu, A. Zeng, J. Ling, H. Cao, K. Li, J. Chen, X. Huang, F. Yan, Y. Chen, et al., “Grounded SAM: Assembling Open-World Models for Diverse Visual Tasks,” arXiv preprint arXiv:2401.14159, 2024.

\bibitem{zhao2023fast} X. Zhao, W. Ding, Y. An, Y. Du, T. Yu, M. Li, M. Tang, and J. Wang,
”Fast Segment Anything,” arXiv preprint arXiv:2306.12156, 2023.

\bibitem{redmon2023yoloworld}
A. Redmon, "YOLO World: Beyond Real-time Object Detection," 
in \textit{Proceedings of the IEEE Conference on Computer Vision and Pattern Recognition (CVPR)}, pp. 5897–5906, June 2023.

\bibitem{krizhevsky2012imagenet}
A. Krizhevsky, I. Sutskever, and G. Hinton,
"ImageNet Classification with Deep Convolutional Neural Networks," 
in \textit{Proceedings of the 25th International Conference on Neural Information Processing Systems (NeurIPS 2012)}, pp. 1097–1105, 2012.

\bibitem{simonyan2014very}
K. Simonyan and A. Zisserman,
"Very Deep Convolutional Networks for Large-Scale Image Recognition," 
in \textit{Proceedings of the 2015 International Conference on Learning Representations (ICLR)}, 2015.

\bibitem{he2016deep}
K. He, X. Zhang, S. Ren, and J. Sun,
"Deep Residual Learning for Image Recognition," 
in \textit{Proceedings of the 2016 IEEE Conference on Computer Vision and Pattern Recognition (CVPR)}, pp. 770–778, June 2016.

\bibitem{ren2015faster}
S. Ren, K. He, R. Girshick, and J. Sun,
"Faster R-CNN: Towards Real-Time Object Detection with Region Proposal Networks," 
in \textit{Proceedings of the 2015 IEEE International Conference on Computer Vision (ICCV)}, pp. 91–99, December 2015.

\bibitem{redmon2016you}
J. Redmon, S. Divvala, R. Girshick, and A. Farhadi,
"You Only Look Once: Unified, Real-Time Object Detection," 
in \textit{Proceedings of the 2016 IEEE Conference on Computer Vision and Pattern Recognition (CVPR)}, pp. 779–788, June 2016.

\bibitem{dosovitskiy2021image}
A. Dosovitskiy, L. Beyer, A. Kolesnikov, D. Weissenborn, X. Zhai, T. Unterthiner, M. Dehghani, M. Minderer, G. Heigold, S. Gelly, J. Uszkoreit, and N. Houlsby,
"An Image is Worth 16x16 Words: Transformers for Image Recognition at Scale,"
in \textit{Proceedings of the International Conference on Learning Representations (ICLR)}, 2021.

\bibitem{yu2022glip}
L. Yu, T. Liu, Y. Chen, H. Li, X. Wang, and J. Chen,
"GLIP: Generalized Local Image-Text Pretraining for Object Detection," 
in \textit{Proceedings of the 2022 IEEE/CVF Conference on Computer Vision and Pattern Recognition (CVPR)}, pp. 12031–12041, June 2022.

\bibitem{caron2021emerging}
M. Caron, H. Touvron, I. Misra, H. Jégou, J. Mairal, P. Bojanowski, and A. Joulin,
"Emerging Properties in Self-Supervised Vision Transformers,"
in \textit{Proceedings of the IEEE/CVF International Conference on Computer Vision (ICCV)}, 2021, pp. 9650-9660.

\bibitem{li2023grounding}
Y. Li, X. Zhao, S. Xu, Y. Wang, C. Li, Z. Lin, and L. Li,
"Grounding DINO: Self-Supervised Learning for Visual Grounding," 
in \textit{Proceedings of the 2023 IEEE/CVF Conference on Computer Vision and Pattern Recognition (CVPR)}, pp. 13057–13067, June 2023.

\bibitem{selvaraju2017grad}
R. R. Selvaraju, M. Cogswell, A. Das, R. Vedantam, D. Parikh, and D. Batra,
"Grad-CAM: Visual Explanations from Deep Networks via Gradient-Based Localization,"
in \textit{Proceedings of the 2017 IEEE International Conference on Computer Vision (ICCV)}, 2017, pp. 618-626.

\bibitem{paszke2019pytorch}
A. Paszke, S. Gross, F. Massa, A. Lerer, J. Bradbury, G. Chanan, T. Killeen, Z. Lin, N. Gimelshein, L. Antiga, A. Desmaison, A. Köpf, E. Yang, Z. DeVito, M. Raison, A. Tejani, S. Chilamkurthy, B. Steiner, L. Fang, J. Bai, and S. Chintala,
"PyTorch: An Imperative Style, High-Performance Deep Learning Library," 
in \textit{Advances in Neural Information Processing Systems 32 (NeurIPS 2019)}, Vancouver, Canada, 2019.

\bibitem{radford2020language}
A. Radford, J. Wu, D. Amodei, and I. Sutskever,
"Language Models are Few-Shot Learners," 
in \textit{Proceedings of the 2020 Conference on Neural Information Processing Systems (NeurIPS)}, December 2020.

\end{thebibliography}
\end{document}